\documentclass[11pt]{acmart}
\usepackage{rotating} 
\usepackage{multirow}
\usepackage{bigstrut}

\AtBeginDocument{%
  \providecommand\BibTeX{{%
    \normalfont B\kern-0.5em{\scshape i\kern-0.25em b}\kern-0.8em\TeX}}}

\setcopyright{acmcopyright}
\copyrightyear{2018}
\acmYear{2018}
\acmDOI{10.1145/1122445.1122456}

\acmConference[Woodstock '18]{Woodstock '18: ACM Symposium on Neural
  Gaze Detection}{June 03--05, 2018}{Woodstock, NY}
\acmBooktitle{Woodstock '18: ACM Symposium on Neural Gaze Detection,
  June 03--05, 2018, Woodstock, NY}
\acmPrice{15.00}
\acmISBN{978-1-4503-XXXX-X/18/06}

\begin{document}

\title{A Comprehensive Survey on Word Representation Models: From Classical to State-Of-The-Art Word Representation Language Models}

\author{Usman Naseem}
\email{usman.naseem@sydney.edu.au}
\authornotemark[1]
\affiliation{%
  \institution{School of Computer Science, The University of Sydney, Australia}
}

\author{Imran Razzak}
\affiliation{%
  \institution{School of Information Technology, Deakin University}
  \country{Australia}}
\email{larst@affiliation.org}

\author{Shah Khalid Khan}
\affiliation{%
  \institution{School of Engineering, RMIT University}
  \country{Australia}
}

\author{Mukesh Prasad}
\affiliation{%
 \institution{School of Computer Science, University of Technology Sydney}
 \country{Australia}}






\renewcommand{\shortauthors}{Naseem U, et al.}

\begin{abstract}

Word representation has always been an important research area in the history of natural language processing (NLP). Understanding such complex text data is imperative, given that it is rich in information and can be used widely across various applications. In this survey, we explore different word representation models and its power of expression, from the classical to modern-day state-of-the-art word representation language models (LMS). We describe a variety of text representation methods, and model designs have blossomed in the context of NLP, including SOTA LMs. These models can transform large volumes of text into effective vector representations capturing the same semantic information. Further, such representations can be utilized by various machine learning (ML) algorithms for a variety of NLP related tasks. In the end, this survey briefly discusses the commonly used ML and DL based classifiers, evaluation metrics and the applications of these word embeddings in different NLP tasks.


\end{abstract}


\maketitle

\section{Introduction}
Text-based data is increasing at a rapid rate, where the low quality of the unstructured text is growing rapidly than structured text. Textual data is very common in many different domains whether social media, online forums, published articles or clinical notes for patients and online reviews given online where people express their opinions and sentiments to some products or businesses \cite{Hu2012}.

Text data is a rich source of getting information and gives more opportunity to explore valuable insights which can not be achieved from quantitative data \cite{Tan99textmining:}. The main aim of different NLP methods is to get a human-like understanding of the text. It helps to examine the vast amount of unstructured and low-quality text and discover appropriate insights. Couple with ML, it can formulate different models for the classification of low-quality text to give labels or obtain information based on prior training. For instance; researchers in the past focused on mining the opinion and sentiments of users about a product, restaurant and movie reviews etc. to predict the sentiment of users. Over the years text has been used in various
applications such as email filtering  \cite{25DBLP:journals/corr/cs-CL-0109015}, Irony and sarcasm detection~\cite{naseem2020towards} document organization~\cite{Hammouda:2004:EPD:1018032.1018338}, sentiment and opinion mining prediction \cite{hyb10.1007/978-3-030-35288-2_31,naseem2020transformer}, hate speech detection \cite{hateIJST146538,naseem2019deep,naseem2019dice,naseem2019abusive}, question answering~\cite{53article}, content
mining~\cite{aggarwal2013data}, biomedical text mining~\cite{naseem2020bioalbert,naseem2020biomedical} and many more.



\begin{figure*}[!htpb]
\hspace{2.5cm}
\centering
\includegraphics[width=0.95\textwidth]{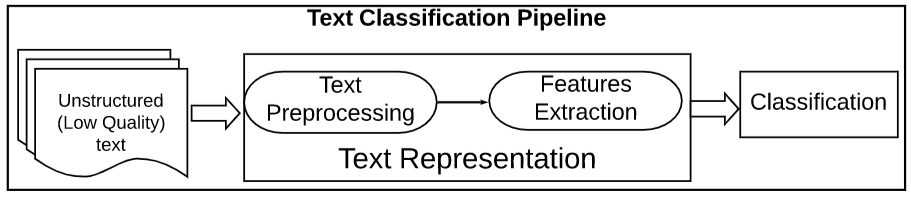}
\caption{Text Classification Pipeline}
\label{fig1}
\end{figure*}

However, being unstructured content, it adds complexity to the model,
deciphers automatically or uses in conjunction with traditional features for a ML framework [57]. Moreover, even
though large of volumes of text information is widely available and can be leveraged for interesting applications, it is rife with
problems. Like most data, it suffers from traditional problems
such as class imbalance and lack of class labels, but besides, there
are some inherent issues with text information. Apart from being
unstructured, text mining and representation learning become
more challenging due to the following discussed factors.

The language on social media is unstructured and informal. Social media users express their emotions and write in different ways, use abbreviations, punctuations, emoticons, slangs and often use URLs. These language imperfections may cause noise and is a challenging task to handle by applying appropriate pre-processing techniques. Besides, understanding semantics, syntactical information and context is important for text analysis~\cite{naseem_dice,naseem2020hybrid}.


Much research has been dedicated to addressing each of these concerns individually. However, in this survey, we focus
on how text can be represented as numeric$\backslash$ continuous vectors for
easier representation, understanding and applicability to traditional
machine-learning frameworks. Text may be seen as a collection of
entities such as documents, sentences, words or characters and
most algorithms leverage the implicit relationship between these
entities to infer them as vectors.

Over the years, many methods and algorithms have been used to
infer vectors from text be at character, word, sentence or document
level. All the methods are aimed at better quantifying the richness
in the information and making them more suitable for machine
learning frameworks such as to perform clustering, dimensionality
reduction or text classification. In this survey, we study how text
representation methods have evolved from manually selecting the features called feature engineering to more SOTA representational learning methods which leverage neural networks to
discover relevant embeddings.

In any NLP task, first, we should have data which we are interested in analyzing. The next step is to represent the raw unstructured data in a form that ML classification algorithms can understand. Text representation is divided into main two parts: i) Text pre-processing and, ii)  Features Extraction and then classify the learned representations using an appropriate classifier \citep{surveyDBLP:journals/corr/abs-1904-08067,surveyAggarwal2012ASO}.

\textbf{Contribution and Organization} In this paper, we present a comprehensive study of various text representation methods starting
from the bag of words approach to more SOTA representational learning methods. We describe various commonly used text
representation methods and their variations and discuss various
text mining applications they have been used in. We conclude with
a discussion about the future of text representation based on our
findings. We would like to note that this paper, strictly focuses
on the representation of text for low-quality text Classification and
therefore uses content, data and text interchangeably.



Below, first, we briefly discuss different steps in text classification pipeline illustrated in Fig.~\ref{fig1}, followed by the details of each step in next sections.



\begin{enumerate}

    \item \textbf{ Unstructured (Low Quality) Text:} 
    Unstructured (Low Quality) text is a form of written text which requires metadata and cannot easily be listed or classified.  Usually, it is the information generated by users on social media postings, documents, email or messages. Raw text is scattered and sparse with less number of features and does not give sufficient word co-occurrence information. It is an important origin of information for businesses, research institutes and monitoring agencies. Often companies mine it for getting the data to improve their marketing strategies and achieve an edge in the marketplace. It plays a big part in predictive analytics and in analysing sentiments of users to find-out the overall opinion of customers. It helps to discover unique insights by revealing hidden information, discovering trends and recognising relationships between irrelevant bits of the data \citep{Haddi2013TheRO,Uysal2014TheIO}.

\item \textbf{Text Representation}: For text classification, the text should be converted into a form which the computer can understand. First, we need to improve the quality of raw and unstructured text and then extract features from it before classification. Both of these steps are briefly discussed below.

\begin{itemize}
    \item \textbf{Text pre-processing:} pre-processing is the crucial step, especially in the classification of short text. pre-processing techniques are valuable techniques for decreasing the data adequacy, sparsity and helps to improve the low quality of text especially in the case of short text where everyone writes in their style, and use emoticons, abbreviations, make spelling mistakes and use URLs etc. A proper combination of common and advance pre-processing techniques can help to learn good text representation \citep{bao:10.1007/978-3-319-09339-0_62,Singh2016RoleOT}. pre-processing techniques analysed in our study are briefly discussed in section \ref{preproc:tech}. 
    
    \item \textbf{ Features Extraction:} Features extractions of the data is the critical step for machines to classify and understand the data like humans. It is the process of transforming raw data into numeric data which machines can understand. Usually, this feature extraction step of transforming a raw data is called a features vector. Extracting robust word representations is not so easy without having a considerable amount of corpus due to diversity of expressing sentiments, emotions and intentions in the English language. However, due to social media platforms, researchers now have access to get an enormous amount of data. However, assigning labels to this massive amount of data collected from social media platforms is not an easy job. To make this annotation process easy, researchers initially worked on finding a sign of sentiment and emotion within the content of the text like emoticons and hashtags \citep{Suttles:2013:DSE:2458308.2458320,Wang2012HarnessingT,surveyDBLP:journals/corr/abs-1904-08067}. Some of the famous classical and current feature extraction algorithms are briefly discussed in section \ref{feature:tech}.

\end{itemize}

\item \textbf{Classification}:
Selecting the best classifier is the essential part of text classification pipeline.  It is hard to find out the most effective and adequate classifier for text classification task without understanding theoretically and conceptually each algorithm. Since the scope of this paper is only restricted to present different text representation methods so we will not discuss different text classification algorithms in detail. These classifiers include famous traditional ML algorithms of text classification such as Logistic Regression which is used in many data mining areas \citep{17article,18Chen2017ACS}, Naive Bayes which is computationally not expensive and works well with less amount of memory \citep{19Larson:2010:IIR:1753126.1753129}, K-nearest Neighbour which is non-parametric methods and Support Vector Machine is a famous classifier which has been widely used in many different areas earlier.  Then tree-based algorithms like random forest and decision tree are discussed followed by deep learning (DL)-based classifiers which are a collection of methods and approaches motivated by the working mechanism of the human brain. These methods utilise the extensive amounts of training input data to achieve the high quality of semantically rich text representations which can be given as input to different ML methods which can make better predictions \citep{Korde2012TEXTCA,surveyDBLP:journals/corr/abs-1904-08067}.

\end{enumerate}

\section{Text Pre-processing } \label{preproc:tech}

Text datasets contain a lot of unwanted words such as stop-words, punctuation, incorrect spellings, slangs, etc. This unwanted noise and words may have an negative effect on the performance of the classification task. Below first, we present the preliminaries where we discuss different methods and techniques related to text pre-processing and cleaning, followed by some literature review where researchers analyzed the effects of text pre-processing techniques. 

\subsection{Preliminaries related to Text Pre-processing}

\begin{itemize}

    \item \textbf{Tokenization} A process of transforming a text (sentence) into tokens or words is known as tokenization. Documents can be tokenized into sentences, whereas sentences can be converted into tokens. In tokenization, a sequence to text is divided into the words, symbols, phrases or tokens \citep{Balazs2016OpinionMA}. The prime objective of tokenization is to find out the words in a sentence.  Usually, tokenization is applied as a first and standard pre-processing step in any NLP task.\citep{token:inproceedings}
    
    \item \textbf{Removal of  Noise, URLs, Hashtag  and User-mentions} Unwanted strings and Unicode are considered as leftover during the crawling process, which is not useful for the machines and creates noise in the data. Also, almost all of tweets messages posted by users, contains URLs to provide extra information, User-mention/tags ($\alpha$) and use hashtag symbol $ "\#" $ to associate their tweet message with some particular topic and can also express their sentiments in tweets by using hashtags. These give extra information which is useful for human beings, but it does not provide any information to machines and considered as noise which needs to be handled. Researchers have presented different techniques to handle this extra information provided by users such as in the case of URLs; it is replaced with tags \citep{Agarwal2011SentimentAO} whereas User-mentions ($\alpha$) are removed \citep{bermingham-smeaton-2011-using,Khan:2014:TTO:2566999.2567126}
    
    



    \item \textbf{Word Segmentation} Word segmentation is the process of separating the phrases, content and keywords used in the hashtag. Moreover, this step can help in understanding and classifying the content of tweets easily for machines without any human intervention. As mentioned earlier, Twitter users use \# (hashtags) in almost all tweets to associate their tweets with some particular topic. The phrase or keyword starting with \# is known as hashtags. Various techniques are presented in the literature for word segmentation in \citep{segmentarticle,seg2inproceedings}.
 


    \item \textbf{Replacing Emoticons and Emojis} Twitter users use many different emoticons and emojis such as:), :(, etc. to express their sentiments and opinions. So it is important to capture this useful information to classify the tweets correctly. There are few tokenizers available which can capture few expressions and emotions and replace them with their associated meanings  \citep{Gimpel_part-of-speechtagging}.

    \item \textbf{Replacement of abbreviation and slang} Character limitations of Twitter enforce online users to use abbreviations, short words and slangs in their posts online. An abbreviation is a short or acronym of a word such as MIA which stands for missing in action. In contrast, slang is an informal way of expressing thoughts or meanings which is sometimes restricted to some particular group of people, context and considered as informal. So it is crucial to handle such kind of informal nature of text by replacing them to their actual meaning to get better performance without losing information. Researchers have proposed different methods to handle this kind of issue in a text, but the most useful technique is to convert them to an actual word which is easy for a machine to understand \citep{Kouloumpis2011TwitterSA, Mullen2006159}.

%
    \item \textbf{Replacing elongated characters}
    Social media users, sometimes intentionally use elongated words in which they purposely write or add more characters repeatedly more times, such as loooovvveee, greeeeat. Thus, it is important to deal with these words and change them to their base word so that classifier does not treat them different words. In our experiments, we replaced elongated words to their original base words. Detection and Replacement of elongated words have been studied by \cite{mohammad} and \cite{Balahur2013SentimentAI}.

    \item \textbf{Correction of Spelling mistakes} Incorrect spellings and grammatical mistakes are very commonly present in the text, especially in the case of social media platforms, especially on Twitter and Facebook. Correction of spelling and grammatical mistakes helps in reducing the same words written indifferently. Textblob is one the library which can be used for this purpose. Norvig's spell correction\footnote{http://norvig.com/spell-correct.html} method is also widely used to correct spelling mistakes.

    \item \textbf{Expanding Contractions} A contraction is a shortened form of the words which is widely being used by online users. An apostrophe is used in the place of the missing letter(s). Because we want to standardize the text for machines to process easily so, in the removal of contractions, shortened words are expanded to their original root /base words. For example, words like how is, I'm, can't and don't are the contractions for words how is, I am, cannot and do not respectively. In the study conducted by  \cite{Boia2013AI}, contractions were replaced with their original words or by the relevant word. If contractions are not replaced, then the tokenization step will create tokens of the word "can't" into "can" "t".

    \item \textbf{Removing Punctuations} Social media users use different punctuations to express their sentiments and emotions, which may are useful for humans but not all much useful for machines for the classification of short texts. So removal of punctuation is common practice in classification tasks such as sentiment analysis. However, sometimes some punctuation symbols like "!" and "?" shows/denotes the sentiments. Its common practice to remove punctuation. \citep{Lin:2009:JSM:1645953.1646003}. whereas, replacing question mark or sign of exclamation with tags has also been studied by  \cite{Balahur2013SentimentAI}. 

    \item \textbf{Removing Numbers} Text corpus usually contains unwanted numbers which are useful for human beings to understand but not much use for machines which makes lowers the results of the classification task. The simple and standard method is to remove them \citep{he-etal-2011-automatically,numberinproceedings}. However, we could lose some useful information if we remove them before transforming slang and abbreviation into their actual words. For example, words like "2maro", "4 u", "gr8", etc. should be first converted to actual words, and then we can proceed with this pre-processing step.

    \item \textbf{Lower-casing all words} A sentence in a corpus has many different words with capitalization. This step of pre-processing helps to avoid different copies of the same words. This diversity of capitalization within the corpus can cause a problem during the classification task and lower the performance. Changing each capital letters into a lower case is the most common method to handle this issue in text data. Although, this pre-processing technique projects all tokens in a corpus under the one feature space also causes a bunch of problems in the interpretation of some words like "US" in the raw corpus. The word "US "could be pronoun and a country name as well, so converting it to a lower case in all cases can be problematic. The study conducted by  \cite{Santos2014DeepCN} has lower-cased words in corpus to get clean words.

    \item \textbf{Removing Stop-words} In-text classification task, there are many words which do not have critical significance and are present in high frequency in a text. It means the words which does not help to improve the performance because they do not have much information for the sentiment classification task, so it is recommended to remove stop words before feature selection step. Words like (a, the, is, and, am, are, etc.). A popular and straightforward method to handle with such words is to remove them. There are different stop-word libraries available such as  NLTK, scikit-learn and spaCy.

    \item \textbf{Stemming} One word can turn up in many different forms, whereas the semantic meaning of those words is still the same. Stemming is the techniques to replace and remove the suffixes and affixes to get the root, base or stem word. The importance of stemming was studied by  \cite{mejovainproceedings}. There are several types of stemming algorithms which helps to consolidate different forms of words into the same feature space such as  Porter Stemmer, Lancaster stemmer and Snowball stemmers etc. Feature reduction can be achieved by utilizing the stemming technique.

    \item \textbf{Lemmatization} The purpose of the lemmatization is the same as stemming, which is to cut down the words to it's base or root words. However, in lemmatization inflection of words are not just chopped off, but it uses lexical knowledge to transform words into its base forms. There are many libraries available which help to do this lemmatization technique. Few of the famous ones are  NLTK (Wordnet lemmatizer), genism, Stanford CoreNLP, spaCy and TextBlob etc.

    \item \textbf{Part of Speech (POS) Tagging} The purpose of Pat of speech (POS) tagging is to assign part of speech to text. It clubs together with the words which have the same grammatical with words together.

    \item \textbf{Handling Negations} For humans, it is simple to get the context if there is any negation present in the sentence, but for machines sometimes it does not help to capture and classify accurately so handling a negation can be a challenging task in the case of word-level text analysis. Replacing negation words with the prefix 'NEG\_' has been studied by  \cite{NegDBLP:journals/corr/abs-1305-6143}. Similarly, handling negations with antonym has been studied by  \cite{Perkins:2010:PTP:1952104}.
    
\end{itemize}


\subsection{Related work on text pre-processing methods}

Text pre-processing plays a significant role in text classification. Many researchers in the past have made efforts to understand the effectiveness of different pre-processing techniques and their contribution to text classification tasks. Below we present some studies conducted on the effects of pre-processing techniques on text classification tasks.

Bao et al.~\cite{bao:10.1007/978-3-319-09339-0_62} study showed the effect of pre-processing techniques on Twitter analysis task. Uni-gram and bi-grams features were fed to Liblinear classifier for the classification. They showed in their study that reservation of URL features, the transformation of negation (negated words) and normalization of repeated tokens have a positive effect on classification results whereas lemmatization and stemming have a negative effect on classification results. Singh and Kumari~\cite{Singh2016RoleOT} showed the impact of pre-processing on Twitter dataset full of abbreviations, slangs, acronyms for the sentiment classification task. In their study, they showed the importance and significance of slang and correction of spelling mistakes and used Support Vector Machine (SVM) classifier to study the role of pre-processing for sentiment classification. Haddi et al.~\cite{Haddi2013TheRO} also explored the effect of text pre-processing on movie review dataset. The experimental shows that pre-processing methods like the transformation of text such as changing abbreviations to actual words and removal of stop word, special characters and handling of negation with the prefix ‘NOT’ and stemming can significantly improve the classification performance. The SVM classifier was used in their experiments — the study conducted by Uysal and Gunal. \cite{Uysal2014TheIO} to analyze the role of pre-processing on two different languages for sentiment classification was presented. They employed SVM classifier in their studies and showed that performance is improved by selecting an appropriate combination of different techniques such as removal of stop words, the lower casing of text, tokenization and stemming. They concluded that researchers should choose all possible combinations carefully because the inappropriate combination may result in degradation of performance.
Similarly,  Jianqiang and Xiaolin~\cite{gui:Jianqiang2017ComparisonRO} studied the role of six different pre-processing techniques on five datasets in their study, where they used four different classifiers. Their experimental results show that replacing acronyms (abbreviations) with actual words and negations improved the sentiment classification, whereas removing stop-words, special characters, and URLs have an adverse influence on the results of sentiment classification. Role of text pre-processing to reduce the sparsity issue in Twitter sentiment classification is studied by Said et al.~\cite{preSaif2013EvaluationDF}. Experimental results demonstrate that choosing a combination of appropriate pre-processing methods can decrease the sparsity and enhance the classification results. Agarwal et al.~\cite{Agarwal2011SentimentAO} proposed novel tweets pre-processing approaches in their studies.  They replaced URL, user mentions, repeated characters and negated words with different tags and removed hashtags. Classification results were improved by their proposed pre-processing methods. In other studies presented by Saloot et al.~\cite{Saloot2015TowardTN} and Takeda and Takefuiji~\cite{Yamada2015EnhancingNE} in the natural language workshop which focuses on noise user-generated text\footnote{http://noisy-text.github.io/}. Noisy nature of Twitter messages is reduced/decreased by normalizing tweets using a maximum entropy model and entity linking. Recently, Symeonidis et al.~\cite{SYMEONIDIS2018298} presented the comparative analysis of different techniques on two datasets for Twitter sentiment analysis. In their study, they studied the effect of each technique on four traditional ML-based classifiers and one neural network-based classifier with only TFIDF (unigram) for words representation method. Their study showed that pre-processing technique such as removing numbers, lemmatization and expanding contractions to base words performs better, whereas removing punctuation does not perform well in the classification task. Their study also presented the interactions of the limited number of different techniques with others and showed that techniques which perform well when interacted with others. However, no work has been done the recommendation of pre-processing techniques to improve the quality of the text.


\section{Feature Extraction methods } \label{feature:tech}

In this section, we discuss various popularly used feature extraction models. Different researchers in the past have proposed different features of extraction models to address the problem of loosing
syntactic and semantic relationships between words. These methods, along with the literature review where different methods have been adopted for different NLP related tasks. First, we present some classical models, followed by some famous representation learning models.


\subsection{Classical Models}\label{LWR} 
This section presents some of the classical models which were commonly used in earlier days for the text classification task. Frequency of words is the basis of this kind of words representation methods. In these methods, a text is transformed into a vector form which contains the number of the words appearing in a document. First, we give a short description of categorical word representation methods and then weighted word representation methods.


\begin{enumerate}

    \item \textbf{Categorical word representation:} is the simplest way to represent text. In this method, words are represented by a symbolic representation either "1" or "0". One-hot encoding and Bag-of-words (BoW) are the two models which come under categorical word representation methods.  Both are briefly discussed below.
    
    \begin{itemize}
        \item \textbf{One hot encoding:} The most straightforward method of text representation is one hot encoding. In one hot encoding, the dimension is the same amount of terms present in the vocabulary. Every term in vocabulary is represented as a binary variable such as 0 or 1, which means each word is made up of zeros and ones. Index of the corresponding word is marked with 1, whereas all others are marked as zero (0). Each unique word has a unique dimension and will be represented by a 1 in that dimension with 0s everywhere else.
        
        

\item  \textbf{Bag-of-Words (BoW)}: BoW is simply an extension of one-hot encoding. It adds up the one-hot representations of words in the sentence. The BOW method is used in many different areas such as NLP, computer vision (CV), and information retrieval (IR) etc. The matrix of words built using BOW ignore the semantic relationship between words and order of word is also ignored along with the grammar.



\begin{figure}[!htpb]
\centering
\hspace{2cm}
\includegraphics[width=0.35\textwidth]{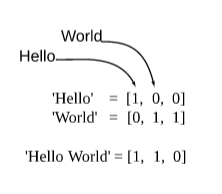}
\caption{An illustration of one-hot encoding and BoW  models} 
\label{bow}
\end{figure}

As stated, BOW is an extension of one-hot encoding, e.g., encodes token in the vocabulary as a 1-hot-encoded vector. As vocabulary may increase to huge numbers, then vocabulary size would increase and thereby, the length of the vectors would increase too. Besides, a large number of "0s" which may result in a sparse matrix, containing no order of text as well as information of the grammar used in the sentences.


An example of "\textit{Hello}" and "\textit{World}" as one-hot encoding and "\textit{Hello World}" as BoW is given in Fig.~\ref{bow}.


    \end{itemize}



\item \textbf{Weighted Word representation:}
Here, we present the common methods for weighted word representations such as Term Frequency (TF) and Term Frequency-Inverse Document Frequency (TF-IDF). These are associated with categorical word representation methods but rather than only counting; weighted models feature numerical representations based on words frequency. Both of them are briefly discussed below.

\begin{itemize}


\item \textbf{Term Frequency (TF)} : Term frequency (TF), is the straightforward method of text feature extraction. TF calculates how often a word occurs in a document. A word can probably appear many times in large documents as compared to small ones. Hence, TF is computed by dividing the length of the document. In other words, TF of a word is computed by dividing it with the total number of words in the document.


    \item \textbf{Term Frequency-Inverse Document Frequency (TF-IDF)}: To cut down the impact of common words such as 'the', 'and' etc. in the corpus, TF-IDF was presented by  \cite{SparckJones:1988:SIT:106765.106782} for text representation. TF here stands for Term frequency which is defined in the above section, and IDF denotes inverse document frequency which is a technique presented to be used with TF to reduce the effect of common words. IDF assigns a more weight to words with either higher or lower frequencies. This combination of TF and IDF method is known as TF-IDF and is represented mathematically by the below equation.



\[ TF-IDF(t,d,D) = TF(t,d)\times \log (\frac{D}{df_{t}}) \]

Where $t$ denotes the terms; $d$ denotes each document; $D$ represents the collection of documents and $df_{t}$ denotes sum of documents with term $t$ in it. TF-IDF is built on the concept of BOW model; therefore, it can not capture the order of words in a document, semantics and syntactical information of words. Hence, TF-IDF is good to use as a lexical level feature.


    \end{itemize}
\end{enumerate}



\subsection{Representation Learning}\label{FL}
Since categorical word representations, models fail to capture syntactic and semantic meaning of the words, and these models suffer the curse of high dimensionality. The shortcomings of these models led the researchers to learn the distributed word representation in low dimension space~\citep{distriNIPS2016_6228}. The limitations of classical feature extraction methods make it use a limited for building a suitable model in ML. Due to this, different models have been presented in the past, which discovers the representations automatically for downstream tasks such as classification. Such methods which discover features itself are called as feature learning or representation learning. 

It is very important because the performance of ML models heavily depends on the representations of the input~\cite{bengio2013representation}. DL-based model, which are good at learning important features itself, is changing traditional feature learning methods. Proper representation can be learned either by utilizing supervised learning methods or unsupervised methods. 

In the area of NLP, unsupervised text representation methods like word embeddings have replaced categorical text representation methods. These word embeddings turned into very efficient representation methods to improve the performance of various downstream tasks due to having a previous knowledge for different ML models. Classical feature learning methods have been replaced by these neural network-based methods due to their good representation learning capacity. Word embedding is a feature learning method where a word from the vocabulary is mapped to $N$ dimensional vector. Many different words embedding algorithms have been presented, and the famous algorithms, for instance, \textit{Word2Vec}, \textit{Glove} and \textit{FastText}, are discussed in this study.


    
First, we briefly present different pre-training methods for learning the word representation of the document. These pre-training methods are classified into three different groups: (i) Supervised learning (SL), (ii) Unsupervised learning (UL), and (iii) Self-supervised learning (SSL). Below  we discuss each of these briefly:    

\begin{enumerate}
    \item \textbf{Supervised learning (SL)} is to learn a feature that translates an input to an output on a basis of input-output pair training data.
    
    \item \textbf{Unsupervised learning (UL)} is to discover some intrinsic information, such as clusters, densities, latent representations, from unlabeled data.
    
\item \textbf{Self-Supervised learning (SSL)} is a hybrid of SL and UL. SSL's learning model is mostly the same as SL, except the training data labels are automated. SSL's main concept is to predict some aspect of the input in some form from other parts. The Masked Language Model (MLM), for instance, is a self-supervised task that tries to predict the masked words in a sentence provided the remaining words.


\end{enumerate}





\subsubsection{Distributed Representations} 

As previously mentioned, hand-crafted features were primarily used to model tasks in natural language before approaches based on neural networks came around and addressed some of the challenges faced by conventional Ml algorithms, such as the dimensionality curse.


\begin{enumerate}

\item     \textbf{Continuous Words Representation (Non-Contextual Embeddings): }
    
Word Embedding is NLP technique in which text from the corpus is mapped as the vectors. In other words, it is a type of learned representation which allows same meaning words to have the same representation. It is the distributed representation of a text (words and documents) which is a significant breakthrough for better performance for NLP related problems. The most significant benefit of word embedding is this that it provides more efficient and expressive representation by keeping the word similarity of context and by low dimensional vectors. Nowadays, word embedding is being used in many different applications like semantic analysis, philology, psychiatry, cognitive science, social science and psychology \citep{elekesarticle}.
    
An automatic feature learning technique in which every token in a vocabulary is indexed into an N dimension vector is known as distributed vectors or Word embedding.
Which follows the distributional hypothesis. According to this, words which are used and appear in the similar contexts tend to assure the same meanings. So these vectors tend to have the attributes of word’s neighbours, and they capture the similarity between words. 
During 1990, several researchers made attempts to lay down the foundation of distributional semantic learning.

Bengio et al.~\cite{Bengio:2003:NPL:944919.944966} presented a model which learned word representations using distributed representation. Authors presented NNLM model which obtains word representations as to the product while training language model (LM). Just like traditional LM, NNLM also uses previous $n-1$ words/tokens to predict the $nth$ word/token. Different word embedding models have been proposed, which makes uni-grams useful and understandable to ML algorithms and usually, these models are used in the first layer in a deep neural network-based model.  These word embedding are pre-trained by predicting a word based on its context without losing semantic and syntactical information. Thus, using these embedding techniques have demonstrated to be helpful in many NLP tasks because It does not lose the order of words and captures the meaning of words (syntactic and semantic information of words). However, the popularity of word representation methods are due to two famous models, Word2Vec \citep{mikolov2013distributed} and GloVe \citep{manning2014stanford}. These famous, along with others, are briefly discussed below.


\begin{itemize}
    \item \textbf{Word2Vec}
    
 Word2vec is words representation model developed by~\cite{mikolov2013distributed}.  This model uses two hidden layers which are used in a shallow neural network to create a vector of each word. The word vectors captured by Continuous Bag of words (CBOW) and Skip-gram models of word2vec are supposed to have semantic and syntactic information of words.  To have a better representation of words, it is recommended to train the corpus with the large corpus. Word2Vec have proved to be useful in many NLP related tasks~\cite{12DBLP:journals/corr/abs-1103-0398}. Word2Vec was developed to build training of embedding more significant, and since then, it has been used as a standard for developing pre-trained word representation. Based on the context, Word2Vec predicts by using one of the two neural network models such as Continuous bag of words (CBOW) and Skip-gram. A predefined length of the window is moved together with the corpus in both models, and the training is done with words inside the window in each step~\cite{Altszyler2016ComparativeSO}. This feature presentation algorithm gives a robust tool for unfolding relationships in the corpus and the similarity between token. For instance, this method would regard the two words such as
“\textit{small}” and “\textit{smaller}” near to each other in the vector space. Fig.\ref{w2v} shows the working principle of both Word2Vec algorithms, \textit{CBOW} and \textit{Skip-Gram}.   
 
 \begin{figure}[!htpb]
\centering
\includegraphics[width=.80\textwidth]{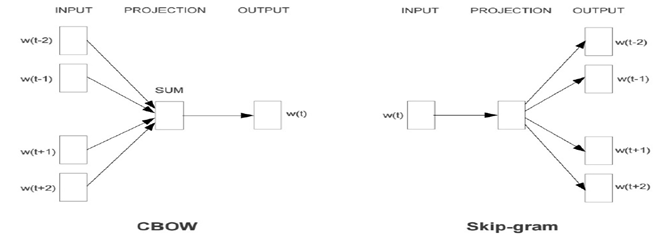}
\caption{Working principle of Word2Vec  \\(Image taken from \citep{mikolov2013distributed})} 
\label{w2v}
\end{figure}

    \begin{itemize}
        \item \textbf{Continuous Bag of words (CBOW):}
        Continuous Bag of words (CBOW) gives words prediction of current work based on its context. CBOW communicates with the neighbouring words in the window. Three layers are used in CBOW process. Context is considered as the first layer whereas the layer which is hidden matches with the estimation of every word from the input to the weight matrix which later on is estimated to the output which is considered as the third layer. The last phase of this method is to correlate the output and the work itself to improve the representation based on the backpropagation of the error gradient. In a Fig.\ref{w2v}, CBOW method predicts the middle word based on its context in skip-gram predicts the context word based on centre word~\cite{NAILI2017340}.
        
    \item \textbf{Skip-Gram:}
    
    Skip-Gram is the reverse of CBOW model; prediction is given based on the central word after the training of context in skip-gram. Input layer correlates with the targeted word, and the output layer corresponds with the context. This model looks for the estimation of the context given the word, unlike CBOW. The last phase of this model is the correlation between output and every word of the context to adjust representation based on back-propagation~\cite{NAILI2017340,elekesarticle}.
    
    Skip-gram is efficient when we have less training data, and not frequent words are well presented. In comparison, CBOW is quicker and performs better with repeated words. To address the issues of learning the final vectors, two algorithms are proposed. First one is negative sampling in which we restrict the sum of output vectors which needs to be updated, so only a sample of the vectors is updated based on a noise distribution which is a probabilistic distribution used in the sample step. Moreover, the other method is Hierarchical softmax which is developed based on the Huffman tree. It is a binary tree which gives all words depending on their counts. Then normalization is done for each step from the root to the target. Negative sampling is efficient when the dimension of vectors is less and works well with repeated words. In comparison, hierarchical softmax works well when we have less frequent words \citep{NAILI2017340}.
    
    \end{itemize}

\item \textbf{Global Vectors (GloVe):}



Word2vec-trained word embedding will better capture the semantics of words and manipulate the connectivity of words. However, Word2vec mainly focuses on the local context window knowledge, whereas the global statistical information is not used well. So the Glove~\cite{manning2014stanford} is presented, which is a famous algorithm based on the global co-occurrence matrix, each element \(X_{ij}\) in the matrix depicts the frequency of the word \(w_{i}\) and the word \(w_{j}\) co-occur in a appropriate context window and is widely used for the text classification task.


GloVe is an expansion of the word2Vec for learning word vectors efficiently where the words prediction is made based on surrounding words. Glove is based on the appearances of a word in the corpus, which is based on two steps. Creation of the co-occurrence matrix from the corpus is the first step, followed by the factorization to get vectors. Like word2Vec, GloVe also provided pre-trained embeddings in a different dimension (100, 200, 300 dimensions) which are trained over the vast corpus. The objective function of GloVe is given below: 


\(J =  \sum_{k,j=1}^{V} f (X_{kj})(w_{k}^{T}{w_{j}^{'}}+b_{k}+b_{j}-\log X_{kj})\)


where;

V : is size of vocabulary,

X : is co-occurrence matrix, 

\(X_{kj}\) is frequency of word k co-occurring with word j,

\(X_{k}\) total number of occurrences of word k in the corpus,

\(P_{kj}\) is the probability of word j occurring within the context of word k,

w is a word embedding of dimension d,

\(w^{'}\) is the the context word embedding of dimension d


    Word representation methods such as Word2vec and GloVe are simple, accurate, and on large data sets, they can learn semantic representations of words. They do not, however, learn embedded words from out-of-vocabulary(OOV) words. Such words can be defined in two ways: words that are
    not included in the current vocabulary and words that do not appear in the current training corpus. To solve these various models are proposed to address this challenge. We briefly describe one of the most famous models below.

\item \textbf{FastText}


   Bojanowski et al.~\cite{bojanowski2016enriching}  proposed FastText and is based on CBOW. When compared with other algorithms, FastText decreases the training time and maintains the performance. Previously mentioned algorithms assign a distinct representation to every word which introduces a limitation, especially in case of languages with sub-word level information/ OOV.

    FastText model solved the issues mentioned above. FastText breaks a word in n-grams instead of full word for feeding into a neural network, which can acquire the relationship between characters and pick up the semantics of words. FastText gives better results by having better word representation primarily in the case of rare words. Facebook has presented pre-trained word embeddings for 294 different languages, trained on Wikipedia using FastText embedding on 300 dimensions and utilized word2Vec skip-gram model with its default parameters \citep{fasstextDBLP:journals/corr/JoulinGBM16}.

\end{itemize}    
Although these models are used to retain syntactic and semantic information of a document, there remains the issue of how to keep the full context-specific representation of a document. Understanding the actual context is required for the most downstream tasks in NLP. Some work recently tried to incorporate the word embedding with the LM to solve the problem of meaning. Below, some of the common context-based models are briefly presented.

  \begin{figure}[!htpb]
\centering
\includegraphics[width=.60\textwidth]{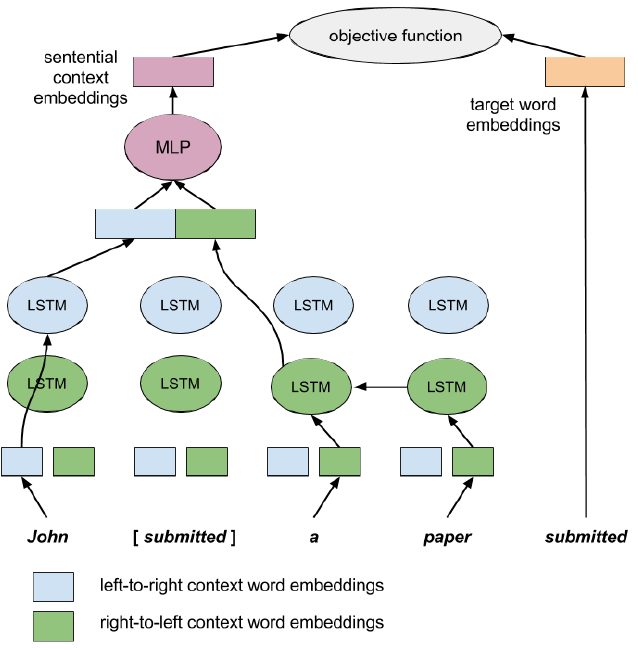}
\caption{Working principle of Context2Vec  \\(Image taken from\citep{melamud-etal-2016-context2vec}) }
\label{c2v}
\end{figure}


\item \textbf{{Contextual word representations}:}


\begin{itemize}

    \item \textbf{Generic Context word representation (Context2Vec):} Generic Context word representation (Context2Vec) was proposed by Melamud et al.~\cite{Melamud2016context2vecLG} in 2016  to generate context-dependent word representations. Their model is based on word2Vec’s CBOW model but replaces its average word representation within a fixed window with better and powerful Bi-directional LSTM neural network. A large text corpus was used to learn neural model which embeds context from a sentence and target words in the same low dimension which later is optimized to reflect the inter-dependencies between target and their entire sentential context as a whole as shown in Fig.~\ref{c2v}.

    
\item \textbf{Contextualized word representations Vectors (CoVe):}

McCann et al.~\cite{coveDBLP:journals/corr/abs-1708-00107}  presented their model contextualized word representations vectors (CoVe) which is based on context2Vec.  They used machine translation to build CoVe instead of the approach used in Word2Vec (skip-gram or CBOW) or Glove (Matric factorization). Their basic approach was to pre-train two-layer BiLSTM for attention sequence to sequence translation, starting from GloVe word vectors and then they took the output of sequence encoder and called it a CoVe, Combine it with GloVe vectors and use in a downstream task-specific mode using transfer learning.


\item \textbf{Embedding from language Models (ELMo)}

Peters et al.~\cite{peterN18-1202} proposed Embedding from Language Models (ELMo), which gives deep contextual word representations. Researchers concur that two problems should be taken into account in a successful word representation model: the dynamic nature of word use in semantics and grammar, and as the language environment evolves, these uses should alter. They therefore introduce a method of deep contextualised word representation to address the two problems above, as seen in Fig.~\ref{elmo}.

  \begin{figure}[!htpb]
\centering
\includegraphics[width=.55\textwidth]{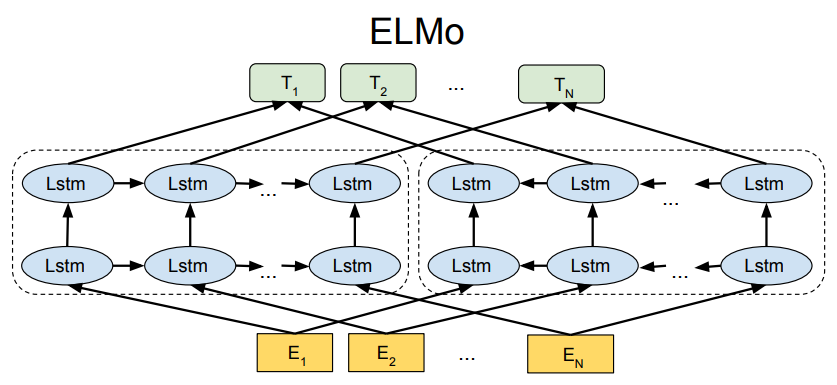}
\caption{Working principle of ELMo  \\(Image taken from\citep{bertDBLP:journals/corr/abs-1810-04805}) }
\label{elmo}
\end{figure}

The final word vectors are learned from bi-directional language model (forward and backward LMs). ELMo uses the linear concatenation of representations learnt from bidirectional language model instead of only just the final layer representations like other contextual word representations. In different sentences, ELMo provides different word representations for the same word. Word representations learned by ELMo are based on the representation learned from Bi-language model (BiLM). The log-likelihood of sentences is used in the training phase of BiLMs both in forward and backward LMs. The final vector is computed after the concatenation of hidden representations from forwarding LM \( \overrightarrow h_{n,j}^{LM}\) and backward LM \( \overleftarrow h_{n,j}^{LM} \), where \(    j = 1,...., L \) and is given below:





\[
 BiLM = \sum_{n=1}^{k} (\log  p(t_n | t_1,.....,t_{n-1};\Theta_x,\overrightarrow \Theta_{LSTM},\Theta_s)  \] 
 \quad \quad   \[ +\log p(t_n|t_{n+1},....,t_n;\Theta_x,\overleftarrow \Theta_{LSTM},\Theta_s)   \]
 

Where the token representation parameters and softmax parameters \(\theta x\) and \(\theta s\) are shared between the forward and backward directions, respectively. And \(\overrightarrow \Theta_{LSTM}\) and \( \overleftarrow \Theta_{LSTM}\) are then forward and backward LSTM parameters respectively.In a downstream task, ELMo extracts the representations learned from BiLM from an intermediate layer and executes a linear combination for each token. BiLM contains \textit{2L+1} set representations as given below.

\[
R_n = ( {X_x^{LM},\overrightarrow h_{n,j}^{LM},\overleftarrow h_{n,j}^{LM} |\quad j = 1,....,L} )  \]
\[ =( h_{n,j}^{LM} | \quad j=0,...,L ) \]

where \(h_{n,0}^{LM}  = {x_n^{LM}}\) is the layer of token and \(h_{n,j}^{LM} = [ \overrightarrow h_{n,j}^{LM},\overleftarrow h_{n,j}^{LM}] \) for each bilstm layer. ELMo is a combination of these characteristics unique to the task where all layers in \textit{M} are flattened to a single vector and are given below:


\begin{equation}\label{eq:2}
ELMo_n^{task} = E(M_n;\Theta^{task}) = \gamma^{task} \sum_{j=0}^{L} s_j^{task}h_{h,j}^{LM}
\end{equation}
Where \( s^{task} \)  are weights which are softmax normalized for the combination of  representations from  different layers and \( \gamma^{task}\)   is a hyper-parameter for optimization and scaling of representations.

Table~\ref{comp} presents the comparison of Classical, non-contextual and contextual (Context2Vec, CoVe and ELMo) LMs with their Pros and cons.

\begin{table}
\centering
\scriptsize
\caption{Comparison of Classical, non-contextual and contextual (Context2Vec, CoVe, ELMo) Word Representation Models}
\label{comp}
\hspace*{-1.5cm}\begin{tabular}{|c|c|c|l|l|} 
\hline
\textbf{Model}                                                        & \textbf{Architecture}  & \textbf{Type}                                                           & \multicolumn{1}{c|}{\textbf{Pros} }                                                                                                                                                                                           & \multicolumn{1}{c|}{\textbf{Cons} }                                                                                                                                                                              \\ 
\hline
\begin{tabular}[c]{@{}c@{}}One Hot Encoding\\ and\\ BoW \end{tabular} & -                      & \multirow{2}{*}{\begin{tabular}[c]{@{}c@{}}Count\\ based \end{tabular}} & \begin{tabular}[c]{@{}l@{}}i) Easy to compute\\ ii) Works with the unknown word\\ iii) Fundamental metric to extract terms \end{tabular}                                                                                      & \begin{tabular}[c]{@{}l@{}}i) It does not capture the semantics syntactic info.\\ ii) Common words effect on the results\\ iii) Can not capture sentiment of words \end{tabular}                                 \\ 
\cline{1-2}\cline{4-5}
\begin{tabular}[c]{@{}c@{}}TF\\ and \\ TF-IDF \end{tabular}           & -                      &                                                                         & \begin{tabular}[c]{@{}l@{}}i) Easy to compute\\ ii) Fundamental metric to extract the descriptive terms \\ iii) Because of IDF, common terms do not impact results\\ \end{tabular}                                            & \begin{tabular}[c]{@{}l@{}}i) It does not capture the semantics syntactic info.\\ ii) Can not capture the sentiment of words \end{tabular}                                                                       \\ 
\hline
Word2Vec                                                              & Log Bilinear           & Prediction based                                                        & \begin{tabular}[c]{@{}l@{}}i) It captures the text semantics syntactic\\ ii) Trained on huge corpus ( Pre-trained) \end{tabular}                                                                                              & \begin{tabular}[c]{@{}l@{}}i) Fails to capture contextual information.\\ ii) It fails to capture OOV words\\ iii) Need huge corpus to learn \end{tabular}                                                        \\ 
\hline
GloVe                                                                 & Log Bilinear           & Count based                                                             & \begin{tabular}[c]{@{}l@{}}i) Enforce vectors in the vector space to identify\\ sub-linear relationships\\ ii) Smaller weight will not affect the training progress\\ for common words pairs such as stop words \end{tabular} & \begin{tabular}[c]{@{}l@{}}i) It fails to capture contextual information \\ ii) Memory utilization for storage\\ iii) It fails to capture OOV words\\ iv) Need huge corpus to learn (Pre-trained) \end{tabular}  \\ 
\hline
FastText                                                              & Log Bilinear           & Prediction based                                                        & \begin{tabular}[c]{@{}l@{}}i) Works for rare words\\ ii) Address OOV words issue. \end{tabular}                                                                                                                               & \begin{tabular}[c]{@{}l@{}}i) It fails to capture contextual information\\ ii) Memory consumption for storage\\ iii) Compared to GloVe and Word2Vec, it is more \\ costly computationally. \end{tabular}         \\ 
\hline
\begin{tabular}[c]{@{}c@{}}Context2Vec\\ CoVe ELMo \end{tabular}      & BiLSTM                 & Prediction based                                                        & i) It solves the contextual information issue                                                                                                                                                                                 & \begin{tabular}[c]{@{}l@{}} i) Improves performance \\ ii) Computationally is more expensive \\ iii) Require another word embedding for all\\LSTM and feed-forward layer \end{tabular}                           \\
\hline
\end{tabular}
\end{table}

\textbf{Summary:} Text representation embeds textual data into a vector space, which significantly affects the performance of downstream learning tasks. Better representation of text is more likely to facilitate better performance if it can efficiently capture intrinsic data attributes. Below we briefly highlight the limitations of categorical and continuous word representation models.

Classical word representation methods like categorical and weighted word representations are the most naive and most straightforward representation of textual data. These legacy word representation models have been used widely in early days for different classification tasks like document classification, Natural language processing (NLP),  information retrieval and computer vision (CV). The categorical word representation models are simple and not difficult to implement but their limitations such as they do not consider capture semantics and syntactic information because they do not consider the order of words and do not consider any relationship between words. Further, the size of the input vector is proportional to vocabulary size, which makes them computationally expensive, which results in poor performance.

    
    
Representation learning methods have helped the research community to build powerful models. However, its drawback is that the features need to be selected manually so to solve this shortcoming there was a need to present some methods which can discover and learn these representations automatically for any downstream task. This automatic extraction of features without human intervention is known as representation learning which has improved results drastically over the past few years in many areas such as image detection, speech recognition, NLP etc. \citep{lecunarticle}. Continuous word representation models like Word2Vec , GloVe  and FastText \citep{mikolov2013distributed,manning2014stanford,fasstextDBLP:journals/corr/JoulinGBM16,DBLP:journals/corr/BojanowskiGJM16} etc. have drastically improved the classification results and overcome shortcomings of categorical representations. It is found that having these continuous word representation of words is more affected as compared to traditional linguistic features because of their ability to capture more semantic and syntactic information of the textual data without losing much information. Despite their success, there are still some limitations which they are not capable of addressing such as they are unable to handle polysemy issues because they assign the same vector to word and ignores its context. Also, models like Word2Vec and GloVe assigns a random vector to a word which they did not encounter during training which means they are unable to handle out of vocabulary (OOV) words which were solved by FastText which breaks words into n-grams. All of these limitations degrades the performance of text classification. Moreover, all of the current SOTA methods do not perform well in the case of the low-quality text.

\begin{table}[!htpb]
\centering
\caption{Gap Analysis of Classic, Non-contextual, Contextual (Context2Vec, Cove and ELMo)LMs}
\label{gapp}
\begin{tabular}{|c|c|c|c|c|}
\hline
\textbf{\begin{tabular}[c]{@{}c@{}}Language\\  Models\end{tabular}} & \textbf{Semantics}     & \textbf{Syntactical}   & \textbf{Context}       & \textbf{\begin{tabular}[c]{@{}c@{}}Out of\\ Vocabulary\end{tabular}} \\ \hline
1-Hot encoding                                                                               & {[}$\times${]}                  & {[}$\times${]}                  & {[}$\times${]}                  & {[}$\times${]}                                                                                                  \\ \hline
BoW                                                                                            & {[}$\times${]}                  & {[}$\times${]}                  & {[}$\times${]}                  & {[}$\times${]}                                                                                         \\ \hline
TF                                                                                             & {[}$\times${]}                  & {[}$\times${]}                  & {[}$\times${]}                  & {[}$\times${]}                                                                                               \\ \hline
TF-IDF                                                                                         & {[}$\times${]}                  & {[}$\times${]}                  & {[}$\times${]}                  & {[}$\times${]}                                                                                              \\ \hline
Word2Vec                                                                                       & [\checkmark] & [\checkmark] & {[}$\times${]}                  & {[}$\times${]}                                                                                          \\ \hline
GloVe                                                                                          & [\checkmark] & [\checkmark] & {[}$\times${]}                  & {[}$\times${]}                                                                                               \\ \hline
FastText                                                                                       & [\checkmark] & [\checkmark] & {[}$\times${]}                  & [\checkmark]                                                                               \\ \hline
Context2Vec                                                                                    & [\checkmark] & [\checkmark] & [\checkmark] & [\checkmark]                                                                              \\ \hline
CoVe                                                                                           & [\checkmark] & [\checkmark] & [\checkmark] & {[}$\times${]}                                                                                            \\ \hline
ELMo                                                                                           & [\checkmark] & [\checkmark] & [\checkmark] & [\checkmark]                                                                            \\ \hline
\end{tabular}%
\end{table}


\item \textbf{Universal Language Model Fine-Tuning (ULMFiT)}

Presented by Jeremy Howard of fast.ai and Sebastian Ruder of the NUI Galway Insight Center, Universal Language Model Fine-tuning (ULMFiT)~\cite{howard-ruder-2018-universal} is basically a method to allow transfer learning and achieve excellent performance for any NLP task, without training models from scratch. ULMFiT proposed two new methods within the network layers, Discriminative Fine-tuning (Discr) and Slanted Triangular Learning Rates (STLR) to enhance the Transfer Learning process. This approach includes fine-tuning a pre-trained LM, trained on the dataset of Wikitext 103, to a new dataset in such a way that it does not neglect what it has learned before. UMFiT was based on the SOTA LM at that time which is LSTM-based model. The architecture and training method, ULMFiT, builds on similar approaches of CoVE and ELMo. In CoVe and ELMo, the encoder layers are frozen. ULMFiT instead describes a way to train all layers, and does so without over-fitting or running into “catastrophic forgetting”, which has been more of a problem for NLP (vs Computer vision) transfer learning in part because NLP models tend to be relatively shallow. Table~\ref{gapp} presents the gap analysis of Classic, Non-contextual, Contextual (Context2Vec, Cove and ELMo)LMs.

  \begin{figure}[!htpb]
\centering
\includegraphics[width=.60\textwidth]{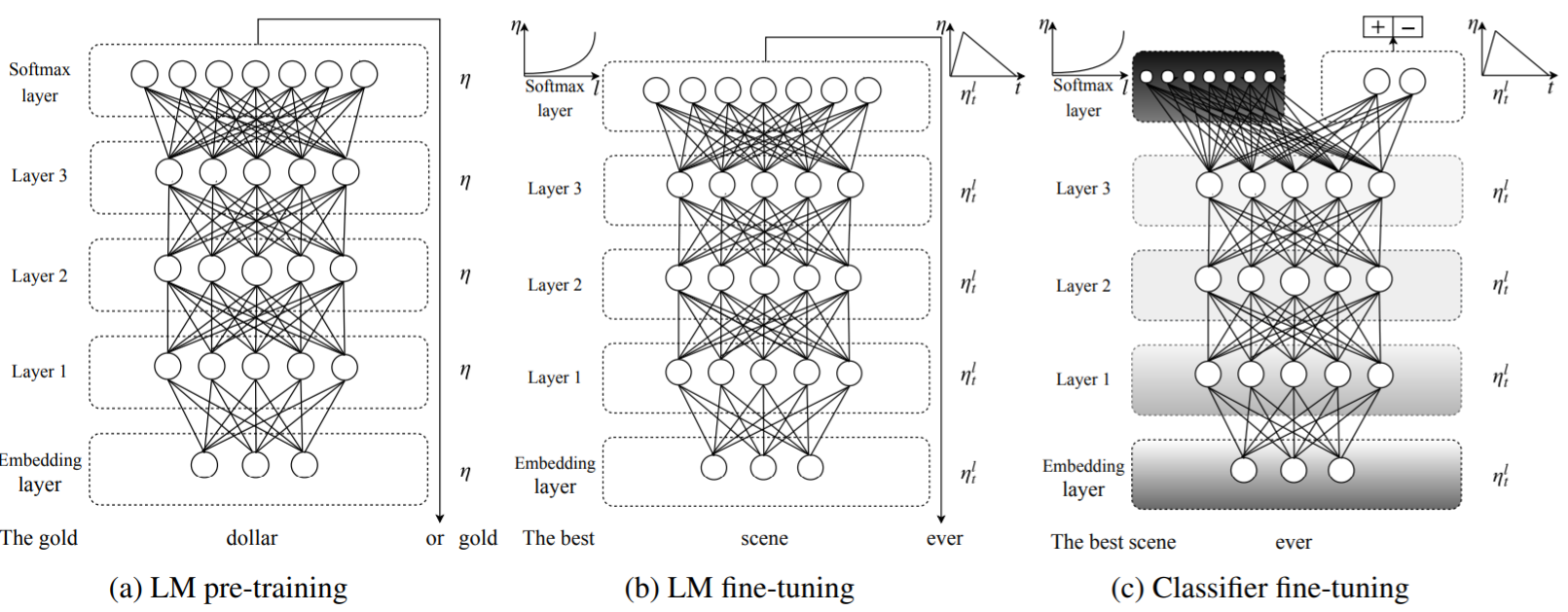}
\caption{Working principle of ULMFiT  \\(Image taken from\citep{howard-ruder-2018-universal}) }
\label{umfit}
\end{figure}

ULMFiT follows three-step to get the good results on downstream tasks, i.e., (i) General LM pre-training, (ii) Target task LM fine-tuning, and (iii) Target task classifier fine-tuning. Three training stages of ULMFiT is shown in Fig.~\ref{umfit}.


The LM pre-training is unsupervised, as the unlabeled text datasets are numerous, the pre-training can be expanded up as much as possible. It still has, however, a reliance on task-customized models. Therefore, the enhancement is only gradual as looking for a better model architecture for each role remains non-trivial until the transformer-based models that are discussed below come into being.

\item \textbf{Transformer-based Pre-trained Language Models}

Transformer~\cite{transssss} has been proven to be more efficient and faster than LSTM or CNN for language modelling, and thus the following advances in this domain will rely on this architecture.


\item \textbf{GPT (OpenAI Transformer):} Generative Pre-Training, GPT~\cite{gpt2noauthororeditor}, is the first Transformer-based pre-trained LM that can effectively manipulate  the semantics of words in terms of context. By learning on a massive set of free text corporas, GPT extends the unsupervised LM to a much larger scale. Unlike ELMo, GPT uses the decoder of the transformer to model the language as it is an auto-regressive model where the model predicts the next word according to its previous context. GPT has shown good performance on many downstream tasks. One drawback of GPT is it's uni-directional, i.e., the model is only trained to predict the future left-to-right context.  The overall model of GPT is shown in Fig.\ref{gpt}.

  \begin{figure}[!htpb]
\centering
\includegraphics[width=.60\textwidth]{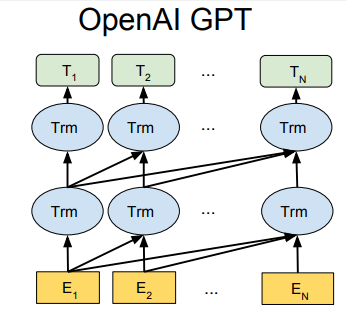}
\caption{Working principle of GPT  \\(Image taken from\citep{gpt2noauthororeditor}) }
\label{gpt}
\end{figure}

\item \textbf{Bidirectional Encoder Representations from Transformers (BERT)}

As seen in Fig.~\ref{bert}, Bidirectional Encoder Representations from Transformers (BERT) is a direct descendant of GPT: train a huge LM on free text and then fine-tune individual tasks without custom network architectures. BERT~\cite{bertDBLP:journals/corr/abs-1810-04805} is another contextualised word representation LM, where the transformer NN uses parallel attention layers rather than sequential recurrence~\cite{transssss}.

  \begin{figure}[!htpb]
\centering
\includegraphics[width=.60\textwidth]{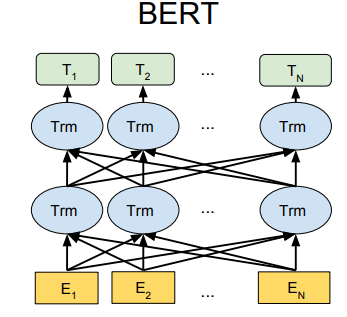}
\caption{Working principle of BERT  \\(Image taken from\citep{bertDBLP:journals/corr/abs-1810-04805}) }
\label{bert}
\end{figure}

Instead of the basic language task, BERT is trained with two tasks to encourage bi-directional prediction and sentence-level understanding. BERT is trained on two unsupervised tasks: (1) a" masked language model (MLM), where 15\% of the tokens are arbitrarily masked (i.e. replaced with the "[MASK]" token), and the model is trained to predict the masked tokens, (2) a "next sentence prediction" (NSP) task, where a pair of sentences are provided to the model and trained to identify when the second one follows the first. This second task is intended to collect additional information that is long-term or pragmatic.

 BERT is trained in the dataset of Books Corpus~\cite{14DBLP:journals/corr/ZhuKZSUTF15} and English Wikipedia text passages. There are two BERT pre-trained model available: BERT-Base and BERT-Large. BERT can be used on un-annotated data or fine-tuned on one's task-specific data straight from the pre-trained model. The publicly accessible pre-trained model and fine-tuning code are available online \footnote{https://github.com/google-research/bert}.
 

\item \textbf{BERT Variants:}

Recent research also explores and strengthens the goal and architecture of
BERT. Some of them are briefly discussed below:

    \item \textbf{GPT2:} The OpenAI team released a scaled-up variant of GPT in 2019 with GPT2~\cite{gpt2noauthororeditor}. It incorporates some slight improvements compared to the previous concerning the position of layer normalisation and residual relations. Overall, there are four distinct GPT2 variants with the smallest being identical to GPT, the medium one being similar in size to BERT-LARGE and the xlarge one being released with 1.5B parameters as the actual GPT2 standard.
    

    \item \textbf{XLNet:} XLNet, also known as  Generalized Auto-regressive Pre-training for Language Understanding~\cite{xlnetDBLP:journals/corr/abs-1906-08237} which proposes a new task to predict the bidirectional context instead of the masked Language task in BERT, it is a permutation language in which we make some permutations of each sentence so the two contexts will be taken into consideration. In order to maintain the position information of the token to be expected, authors employed two-stream self-attention. XLNET was presented to overcome the issue of pre-training fine-tune discrepancy and to include bidirectional contexts simultaneously. 
    
    \item \textbf{RoBERTa:} RoBERTa: A Robustly Optimized BERT Pre-training Approach was implemented in July 2019~\cite{robertDBLP:journals/corr/abs-1907-11692}, it is like a lite version of BERT, but it has fewer parameters and better performance as it removes the training on the sentence classification task. RoBERTa made following changes to the BERT model:  (1) Longer training of the model with larger batches and more data; (2) Eliminating the NSP goal; (3) Longer sequence training; (4) Dynamically
during pre-training, the masked roles will change. All these changes boost the model's efficiency and make it efficient with XLNet 's previous SOTA results.

    \item   \textbf{ALBERT:} Despite this success, BERT has some limitations such as BERT has a huge number of parameters which is the cause for problems like degraded pre-training time, memory management issues and model degradation etc~\cite{albertlan2019albert}. These issues are very well addressed in ALBERT, which is modified based on the architecture of BERT and proposed by Lan et al.~\cite{albertlan2019albert}. In scaling pre-trained models, ALBERT implements two-parameter reduction methods that lift the essential barriers. (i) factorized embedding parameterization - decomposes the big vocabulary embedding matrix to two small matrices, (ii) replaces the NSP loss by SOP loss; and  (iii) cross-layer parameter sharing- stops the parameter from prospering with the network depth. These methods significantly lower the number of parameters used when compared with BERT without significantly affecting the performance of the model, thus increasing parameter-efficiency. An ALBERT configuration is the same as BERT (large) has 18 times less parameters and can be trained about 1.7 times faster. ALBERT establishes new SOTA results while having fewer parameters compared to BERT.

    \item \textbf{Other Models:} Some of the other recently proposed LMs are a cross-lingual LM Pre-training (\textbf{XLM})~\cite{xlmDBLP:journals/corr/abs-1901-07291} from Facebook enhanced BERT for Cross-lingual LM. Two unsupervised training objectives that only include monolingual corporations were introduced by XLM authors: Causal Language Modeling (CLM) and Masked Language Modeling (MLM) and demonstrated that both the CLM and MLM approaches have powerful cross-lingual features that can be used for pre-training models. Similarly, \textbf{StructBERT}~\cite{strucbertwang2019structbert} implemented a structural objective word that randomly permits the reconstruction order of 3-grams and a structural objective sentence that predicts the ordering of two consecutive segments.

    \textbf{DistilBERT:}~\cite{sanh2019distilbert}, a distilled version of BERT, reduces the size of a BERT LM by 40\% while retaining 97\% of its language understanding proficiency and being 60\% quicker. \textbf{MegatronLM}~\cite{megatronshoeybi2019megatronlm}, a scaled-up transform-based model, 24 times larger than BERT, training multi-billion parameter LMs using model parallelism. \textbf{CRTL}~\cite{ctrlkeskar2019ctrl}, A Conditional Transformer Language Model for Controllable Generation, is a 1.63 billion-parameter conditional transformer LM, it is a conditional generative model. Another recently proposed model, \textbf{ERNIE}~\cite{sun2019ernie}, Enhanced representation through knowledge integration, used knowledge masking techniques including entity-level masking and
    phrase-level masking instead of randomly masking tokens. Authors of ERNIE extended their work and presented \textbf{ERNIE 2.0}~\cite{sun2020ernie} further incorporated more pre-training
    tasks, such as semantic closeness and discourse relations. \textbf{SpanBERT}~\cite{joshi2020spanbert} generalized ERNIE to mask random spans, without indicating to external knowledge.

    Other prominent LM includes \textbf{UniLM:}~\cite{dong2019unified}, which uses three objective functions : (i) language modelling (LM), (ii) masked language modelling (MLM), and (iii) sequence-to-sequence language modelling (seq2seq LM), for pre-training a transformer model. In order to monitor what context the prediction conditions are in, UniLM uses special self-attention masks. \textbf{ELECTRA}~\cite{clark2020electra} proposed more better pre-training techniques as compared to BERT. Authors of ELECTRA replaced some of the input tokens with their plausible substitute samples from small generator network rather than corrupting some positions of the inputs with [MASK]. ELECTRA trains a discriminator to determine whether or not each token has been substituted by a generator in the corrupted input that can be used for fine-tuning in downstream tasks. MASS~\cite{song2019mass} is another recently proposed LM. In order to pre-train sequence-to - sequence models, MASS uses masked sequences and adopts an encoder-decoder system and expands the MLM objective. Without pre-training or with other pre-training approaches, MASS makes substantial improvements over baselines on a range of zero / low-resource language generation tasks, including neural machine translation (MT), text summarization and  conversational response generation.

    \item \textbf{Text-to Text Transfer Transformer (T5):}~\cite{raffel2019exploring}, unified natural language understand and generation by transforming the data to the format of text-to-text and apply the framework of an encoder-decoder. In terms of pre-training objectives, architectures, pre-training datasets and transfer techniques, T5 has implemented a novel pre-training corpus and also systematically contrasts previously proposed methods. T5 adopts a text infilling objective, more extended training and multi-task pre-training. For fine-tuning T5 uses the token vocabulary of the decoder as the prediction labels.


    \item \textbf{BART:}\cite{lewis2019bart}: For pre-training sequence-to-sequence models, BART added additional noise functions beyond MLM. First, using an arbitrary noise function, the input sequence is distorted. Then, a transformer network reconstructs the corrupted input. BART explores a broad range of noise functionality, including token functions.
    masking, deletion of tokens, text infilling, rotation of documents and shuffling of words. The best performance 
    is attained by using both sentence shuffling
    and text infilling. BART matches RoBERTa's performance on GLUE and SQuAD and
    attain SOTA results on a number of tasks for generating text.



\begin{table}[!htpb]
\centering
\caption{A comparison of popular Language models.
}
\label{obj}
\resizebox{\textwidth}{!}{%
\begin{tabular}{|c|c|c|c|c|}
\hline
\textbf{LMs} & \textbf{Release Date} & \textbf{Architecture}                                                         & \textbf{Pre-Training   Task} & \textbf{Corpus Used}                                                                            \\ \hline
Word2Vec     & Jan-13                & FCNN                                                                          & -                            &  Google News                                                                 \\ \hline
GloVe        & Oct-14                & FCNN                                                                          & -                            & {\color[HTML]{292929} Common Crawl corpus}                                                      \\ \hline
FastText     & Jul-16                & FCNN                                                                          & -                            & {\color[HTML]{292929} Wikipedia}                                                                \\ \hline
ELMo         & Feb-18                & BiLSTM                                                                        & BiLM                         & Wiki-Text-103                                                                                    \\ \hline
GPT          & Jun-18                & \begin{tabular}[c]{@{}c@{}}Transformer\\  Decoder\end{tabular}                & LM                           & Book-Corpus                                                                                      \\ \hline
GPT-2        & Jun-18                & \begin{tabular}[c]{@{}c@{}}Transformer\\  Decoder\end{tabular}                & LM                           & Web-Text                                                                                         \\ \hline
BERT         & Oct-18                & \begin{tabular}[c]{@{}c@{}}Transformer\\  Encoder\end{tabular}                & MLM \& NSP                   & WikiEn+Book-Corpus                                                                               \\ \hline
RoBERTa      & Jul-19                & \begin{tabular}[c]{@{}c@{}}Transformer\\  Encoder\end{tabular}                & MLM \& NSP                   & \begin{tabular}[c]{@{}c@{}}Book-Corpus + CC-News\\      +Open-Web-Text+ STORIES\end{tabular}          \\ \hline
ALBERT       & Sep-19                & \begin{tabular}[c]{@{}c@{}}Transformer\\  Encoder\end{tabular}                & MLM+SOP                      & same as BERT                                                                                    \\ \hline
XLNet        & Jun-19                & \begin{tabular}[c]{@{}c@{}}Auto-regressive Transformer\\  Encoder\end{tabular} & PLM                          & \begin{tabular}[c]{@{}c@{}}WikiEn+   Book-Corpus+Giga5\\      + Clue-Web + Common Crawl\end{tabular} \\ \hline
ELECTRA      & 2020                  & \begin{tabular}[c]{@{}c@{}}Transformer\\  Encoder\end{tabular}                & RTD+MLM                      & same as XLNet                                                                                   \\ \hline
UniLM        & 2020                  & \begin{tabular}[c]{@{}c@{}}Transformer\\  Encoder\end{tabular}                & MLM+NSP                      & WikiEn + Book-Corpus                                                                               \\ \hline
MASS         & 2020                  & Transformer                                                                   & Seq2Seq MLM                  & *Task-dependent                                                                                 \\ \hline
BART         & 2020                  & Transformer                                                                   & DAE                          & same as RoBERTa                                                                                 \\ \hline
T5           & 2020                  & Transformer                                                                   & Seq2Seq MLM                  & Colossal Clean   Crawled Corpus (C4)                                                            \\ \hline
\end{tabular}
}
\end{table}

\end{itemize}
\end{enumerate}

Although these models were able to solve context issues but are trained on general domain
corpora such as Wikipedia, which limits their applications to specific domains or tasks. To enhance the performance in sub-domains, domain-specific transformer-based models have been proposed. Some of the most famous in the biomedical domain are Sci-BERT~\cite{beltagy2019scibert}, BioBERT~\cite{lee2019biobert} and BioALBERT~\cite{naseem2020bioalbert}. Recently, other domain-specific models such as BERTTweet~\cite{nguyen2020bertweet}, COVID Twitter BERT (CT-BERT)~\cite{muller2020covid} have been trained on datasets from Twitter. Domain-specific models were shown to be useful replacements for LMs trained on general corpus for various downstream tasks. In Table\ref{obj}, we present the architecture, Objective function and dataset used for training in LMs.

\subsection{Related work on Word representation methods}

Below we present some relevant studies where different word representation models have been employed for various text classification tasks. 


Pang et al.~\cite{Pangrel:2002:TUS:1118693.1118704} performed that binary classification task on IMDb dataset and employed unigrams, bigram and POS tags as features. For classification, they used SVM, Maximum entropy and NB classifiers in their study and found out that best results were achieved with unigrams as feature and SVM as classifier. Kwok and Wang~\cite{Kwok:2013:LHD:2891460.2891697} used n-grams features along with NB classifier tweets classification. These legacy based word representation methods such as n-grams, BoW, TF, and TF-IDF have been widely used in different studies for various text classification tasks \cite{Greevy2004AutomaticTC, davidson, Liu:2014:CNB:3094471.3094536,Kouloumpis2011TwitterSA}. These traditional methods for text classification are simple, computationally economical. However, their limitations such as ignoring word order, unable to capture semantic information and high dimensionality etc. restrict their use for efficient text classification tasks.

Later, representation learning methods of learning text representation directly using neural network \cite{Collobert:2008:UAN:1390156.1390177} was adopted, which improved classification results. Word embeddings from continuous word representation models such as Word2Vec and GloVe are the most famous and widely used ones among these methods because of low dimensionality of vectors and capture semantic relationships. Word representation models have also been used for sentence-level classification task by averaging word vectors as feature representation which is utilized later on as input for sentence-level classification \cite{10.1007/978-3-319-19581-0_6}.

Word embeddings which are created based on unigrams and by averaging embeddings are not able to capture the issue of syntactic dependencies like "but" and "negations" can change the complete meaning of a sentence and long dependencies within a sentence \cite{10.1007/978-3-319-19581-0_6}. Sochet et al.~\cite{socherarticle} proposed a recursive neural network which can capture and model long semantic and sentiment dependencies of words and sentence at different stages. The disadvantage of this method is that it depends on parsing, which makes it challenging to use on Twitter related text \cite{foster-etal-2011-news}. A paragraph representation model solved this issue learns word vectors and does not reply on parsing \cite{paraDBLP:journals/corr/LeM14}. Both of these, recursive neural and paragraph representation models have assessed on IMDb dataset used by Pang et al.~\cite{pang2002thumbs}, and both models improved the classification results obtained by using BoW features.


Deep neural network-based methods have also been used for Text classification tasks. Tang et al.~\cite{tang2016sentiment} proposed sentiment specific word representation model, which are achieved from emoticons labelled tweet messages with the help of the neural network. Severyn and Moschitti~\cite{Severyn:2015:TSA:2766462.2767830} presented another neural network-based model where they used Word2Vec to learn embedding. Tweets are presented as a matrix wherein which columns compare with words, thus retaining the position they appear in a tweet. Emoticons annotated data was utilized to pre-train the weights and then trained by hand-annotated from SemEval contestant. The experimental results tell that pretraining step enables for a better initialization of the networks' weights and therefore, has a positive role in classification results. In another study conducted by Fu et al.~\cite{Fu2017CombineHL}, Word2Vec was employed to get word representation which was forwarded to the recursive encoder as an input for text classification. Ren et al.~\cite{Ren:2016:CTS:3015812.3015844} also used Word2Vec to generate word representations and proposed a new model for the Twitter classification task.  Lauren et al.~\cite{laurenpub.1091311082} presented a different document representation model where they used the skip-gram algorithm for generating word representations for the classification of clinical texts.


Due to the limitations and restrictions in a few corpora, pre-trained word embeddings are preferred by researchers as an input of ML models. Qin et al.~\cite{Qin:2016:ECN:2975611.2975670} used pre-trained Word2Vec embeddings and forwarded these word embeddings to CNN. Similarly, Kim~\cite{kim2014convolutional} utilized pre-trained embeddings of Word2Vec and forwarded to CNN neural network, which increased the classification results. Camacho et al.~\cite{CAMACHOCOLLADOS201636} for concept representation in their work. Jianqiang and Xiolin~\cite{DCNN} have initialized word embeddings using pre-trained GloVe embeddings in their DCNN model. Similarly, Wallace~\cite{zhang-wallace-2017-sensitivity} applied GloVe, and Word2Vec pre-trained word embedding in deep neural network-based algorithms and enhanced the classification results. Similarly, a study conducted by  Wang et al.~\cite{wang-etal-2016-attention}, used pre-trained GloVe embeddings as an input to LSTM with attention model for aspect based classification and Liu et al.~\cite{LIU20182287}  employed pre-trained word embeddings Word2Vec for recommending idioms in essay writing. Recently, Ilic et al.~\cite{elmo_ironyDBLP:journals/corr/abs-1809-09795} have used contextual word embeddings (ELMo) for word representation for the detection of sarcasm and irony and shown that using ELMo word representations have improved the classification results. The research community has made limited efforts for solving the above mention limitations of continuous word representation models by proposing different models. For example, for handing OOV words which are not seen in the training and they are assigned UNK token and same vector for all words and degrades results if the number of OOV is large. Different methods to handle  OOV words have been proposed in different studies \cite{dhingDBLP:journals/corr/DhingraLSC17,HerboltDBLP:journals/corr/HerbelotB17,pinterDBLP:journals/corr/PinterGE17} But still these models does not capture the polysemy issues. This issue of words with different meanings (polysemy) is addressed in different models presented by the \cite{neelaDBLP:journals/corr/NeelakantanSPM15, Iacobacci2015SensEmbedLS}. In recent days, researchers has presented more robust models to handle OOV words and polysemy issues \cite{Liu:2015:LCW:2832415.2832428, melamud-etal-2016-context2vec,McCann2017LearnedIT,peterDBLP:journals/corr/abs-1802-05365}.

To handle domain-specific problems different studies have been conducted where researchers used existing knowledge encoded in semantic lexicons to these word embedding to improve the downsides of using the pre-trained embedding which is trained on news data which is usually different from the data we use in our tasks. Some of the models presented are proposed in the following studies which inject external knowledge in existing word embedding and improves the results \cite{faruqiDBLP:journals/corr/FaruquiDJDHS14,conceptDBLP:journals/corr/SpeerCH16,mrksicDBLP:journals/corr/MrksicVSLRGKY17,you2017arXiv170906680Y,NieblerDBLP:journals/corr/Niebler0PH17}. Word embeddings are beneficial in different areas beyond NLP like link prediction, information retrieval and recommendation systems. Ledell et al.~ \cite{wu2017arXiv170903856W} proposed a model which is suitable for many of the applications mentioned above and acted as a  baseline. None of the above mentioned is robust enough and fails to integrate sentiment of words in the representations and does not work well in domain-specific tasks such as sentiment analysis etc.

Studies show that adding sentiment information into conventional word representation models improves performance. To integrate the sentiment information into word embeddings, researchers have proposed different hybrid word representations by changing existing skip-gram model \cite{SSSEtang-etal-2014-learning}.  Tang et al.~\cite{Tang:2016:SEA:2914176.2914206} proposed several hybrid ranking models (HyRank) and developed sentiment embeddings based on C\&W, which considers context and sentiment polarity of tweets. Similarly, several other models are presented, which considers context and sentiment polarity of words for sentiment analysis~\cite{tang2015joint, refineYu:2018:RWE:3180756.3186455,IWVDBLP:journals/corr/abs-1711-08609}. Yu et al.~\cite{refineYu:2018:RWE:3180756.3186455} proposed sentiment embeddings by refining pre-trained embeddings Re(*) using the intensity score of external knowledge resource.  Rezaeinia et al.~\cite{IMVnewDBLP:journals/corr/abs-1711-08609}  proposed improved word vectors (IWV) by combining word embeddings, part of speech (POS) and combination of lexicons for sentiment analysis. Recently, Cambria et al.~\cite{sentic5.Cambria2018SenticNet5D} proposed context embeddings for sentiment analysis by conceptual primitives from text and linked with commonsense concepts and named entities.

Recent studies have used these contextual and transformer-based LMs in their model in various NLP tasks. Furthermore, various studies have been presented which use the domain-specific LMs for different NLP tasks. These hybrid and domain-specific LMs have improved the performance and ability to capture complex word attributes, such as semantics, OOV, context, and syntax, into account in various NLP task.


\section{Classification techniques}



Choosing an appropriate classifier is one of the main steps in the text classification task.  Without having a comprehensive knowledge of every algorithm, we cannot find out the most effective model for the text classification task. Out of many ML algorithms used in text classification, we will present some famous and commonly used classification algorithms. These are used for sentiment classification tasks such as Naïve Bayes (NB), Support vector machine (SVM), logistic regression (LR), Tree-based classifiers like decision tree (DT) and random forest (RF) and neural network-based (DL) algorithms. Table \ref{comp_clas_algo} presents the pros and cons of classification algorithms. 

\subsection{ML based classifiers}

\begin{itemize}
    \item \textbf{Naive Bayes(NB) classifiers}: The Naive Bayes (NB) classifiers are a group of different classification algorithms which are based on Bayes theorem, presented by Thomas Bayes \citep{NB10.2307/2284038}. All Naive Bayes algorithms have the same assumption, i.e., each pair of features being classified is independent of others. The NB classification algorithms are widely used for information retrieval \citep{IRQu2018ImprovedBM} and many text classification tasks \citep{TCPak2010TwitterAA,TC2Melville:2009:SAB:1557019.1557156}. Naive Bayes classifiers are called "Naive" because it considers that every feature is independent of other features in the input. Whereas in reality, words and phrases in the sentences are highly interrelated. The meaning and sentiment depend on the position of words in the sentence, which can change if the position is changed. 
    
    NB classifiers are derived from Bayes theorem which states that given the number of documents (n) to be classified into z classes where \(z \in  \{x_{1},x_{2},....x_{z}\}\) the predicted label out is \(x\in X\).   The Naive Bayes theorem is given as follows:

    \[P (x|y)= \frac{P(y|x)P(x)}{P(y)} \]
    
     Where $y$ denotes a document and $x$ refers to the classes. 
     In simple words, the NB algorithm will take each word in the training data and will calculate the probability of that word being classified. Once the probabilities of every word are calculated, then classifier is read to classify new data by utilizing the prior calculated probabilities during the training phase. Advantages of NB classifiers are; they are scalable,  more suitable when the dimension of input is high, its implementation is simple, less computationally expensive, works well when less training data is available and can often outperform other classification algorithms. Whereas the disadvantages are;  NB classifiers make a solid makes a reliable hypothesis on the shape of data distribution, i.e. any two features are independent given the output class, which gives bad results \citep{SoheilyKhah2017IntrusionDI,Wang_2012}. Another limitation of NB classifiers is due to data scarcity. For any value of the feature, we have to approximate the likelihood value by a frequentist 
     
     \item \textbf{Support vector machine (SVM)}: The support vector machine (SVM) classifiers are one of the famous and common used algorithms used for text classification due to its good performance. SVM is a non-probabilistic binary linear classification algorithm which performs by plotting the training data in multi-dimensional space. Then SVM categories the classes with a hyper-plane. The algorithm will add a new dimension if the classes can not be separated linearly in multi-dimensional space to separate the classes. This process will continue until a training data can be categorized into two different classes. 
     
     
     
     The advantage of SVM classifiers is that results are obtained by using SVM are usually better. The disadvantage of SVM algorithms is that it is not easy to choose a suitable kernel, long training time in case of extensive data and more computational resources are required etc.
\begin{table*}[ht]
\centering
\caption{Comparison of Classification Algorithms}
\label{comp_clas_algo}
\resizebox{\textwidth}{!}{%
\begin{tabular}{|c|l|l|}
\hline
\textbf{Classifiers} & \multicolumn{1}{c|}{\textbf{Pros}}                                                                                                                                                                                        & \multicolumn{1}{c|}{\textbf{Cons}}                                                                                                                                                                                                                                         \\ \hline
NB                   & \begin{tabular}[c]{@{}l@{}}i) Less computational time.\\ ii) Easy to understand \& implement. \\ iii) Can easily be trained less data.\end{tabular}                                                                       & \begin{tabular}[c]{@{}l@{}}i) Relies strongly on the class features independence\\  and does not perform well if the condition is not met.\\ ii) Issue of zero conditional probability for zero\\ frequency  features which makes total probability zero\end{tabular} \\ \hline
SVM                  & \begin{tabular}[c]{@{}l@{}}i) Effective in higher dimension\\ ii) Can model non linear decision boundary\\ iii) Robust to the issue of over-fitting\end{tabular}                                                          & \begin{tabular}[c]{@{}l@{}}i) More computational time for large datasets\\ ii) Kernal selection is difficult\\ iii) Does not perform well in case of overlapped\\ classes\end{tabular}                                                                                       \\ \hline
LR                   & \begin{tabular}[c]{@{}l@{}}i) Easy and Simple to implement.\\ ii) Less computationally expensive\\ iii) Does not need tuning and features \\ to be uniformly distributed\end{tabular}                                     & \begin{tabular}[c]{@{}l@{}}i) Fails in case of non-linear problems\\ ii) Need large datasets\\ iii) Predict results on the basis of independent\\ variables\end{tabular}                                                                                                     \\ \hline
DT                   & \begin{tabular}[c]{@{}l@{}}i) Interpretable and easy to understand\\ ii) Less pre-processing required\\ iii) Fast and almost zero hyper-parameters\\ to tuned\end{tabular}                                                  & \begin{tabular}[c]{@{}l@{}}i) High chances of over fitting\\ ii) Less prediction accuracy as compared to others\\ iii) Complex calculation in large number of classes\end{tabular}                                                                                         \\ \hline
RF                   & \begin{tabular}[c]{@{}l@{}}i) Fast to train, flexible and gives high \\results ii) Less variance than single DT\\ iii) Less pre-processing required\end{tabular}                                                          & \begin{tabular}[c]{@{}l@{}}i) Not easy and simple to interpret\\ ii) Require more computational resources\\ iii) Require more time to predict as compared to\\ others\end{tabular}                                                                                           \\ \hline
DL                   & \begin{tabular}[c]{@{}l@{}}i) Fast predictions once training is complete\\ ii) Works well in case of huge data\\ iii) Flexible architecture, can be utilized for\\ classification  and regression tasks\end{tabular} & \begin{tabular}[c]{@{}l@{}}i) Require a large amount of data\\ ii) Computationally expensive and time-consuming\\ iii) DL based classifiers are like black-box (issue of\\  model interpretability exists)\end{tabular}                                                    \\ \hline
\end{tabular}%
}
\end{table*}
  
    \item \textbf{Logistic Regression (LR) classifier}: Logistic regression (LR) is a statistical model and is one of the earliest techniques used for classification. LR predicts probabilities rather than classes \citep{Fan:2008:LLL:1390681.1442794,Lrdoi:10.1198/004017007000000245} or existence of an event like a win/lose or healthy/sick etc. This can be expanded to model many classes of events like deciding whether an image consists of a cat, duck, cow, etc. Every object being identified in the image would be given a probability between 0 and 1 and the sum adding to one. LR predicts the results based on the set of independent values. However, if the wrong independent values are added, then the model will not predict good results. It works well in the case of categorical results but fails in the case of continuous results. Also, LR wants that every data point to be independent of all others, but if the findings are interlinked to one another, then classifier will not predict good results.

    \item \textbf{Decision Tree (DT) classifier}: Decision tree (DT) was presented by \citet{162Magerman:1995:SDM:981658.981695} and developed by  \citet{163Quinlan:1986:IDT:637962.637969}. It is one of the earliest classification models for text and data mining and is employed successfully in different areas for classification task \citep{16010.2307/2283276}. The main intuition behind this idea was to create tree-based attributes for data points, but the major question is which feature could be a parent and which will be a child's level. DT classifier design contains a root, decision and leaf nodes which denote dataset, carry out the computation and performs classification respectively. During the training phase, the classifier learns the decision need to be executed to separate labelled categories. To classify the unknown instance, the data is passed through the tree. A particular feature from the input text is matched with the fixed which was known during the training stage. The calculation at each decision node compares the chosen features with this fixed feature earlier; the decision relies on whether the feature is more prominent than or less than the fixed which creates two-way division in the tree. The text will eventually go over these decision nodes until it reaches the leaf node that describes it assigned class. 
    
    The advantages of DT classifier are; 
the amount of hyper-parameters which require tuning is nearly zero, easy to describe, can be understood easily by its visualizations whereas the significant disadvantages of DT classifier are; it is sensitive to a minor change in the data \citep{165Giovanelli2017TowardsAA} and have a probability of overfitting \citep{166QUINLAN1987221}, complex computations in case of a large number of class labels and have difficulties with-out-of sample prediction.




    
    \item \textbf{Random Forest (RF) Classifier}: Random forest which is also called an ensembles learning technique for text classification which concentrates on methods to compare the results of several trained models in line to give better classifier and performance than a single model. \citet{Ho:1998:RSM:284980.284986} proposed RF classifier, which is simple to understand and also gives better results in classification. RF classifier is composed of the number of DT classifiers where every tree is trained by a bootstrapped subset of the training text. An arbitrary subset of the characteristics is selected at every decision node, and the model will only examine part of these features. The primary issue with utilizing the single tree is that it has massive variation so that the arrangement of the training data and features can impact its results.

    This classifier is quick to train for textual data but slow in giving predictions when trained \citep{171book}. Performs good with both categorical and continuous variables, can automatically handle missing values, robust to outliers and less affected by noise whereas training a vast number of trees can be computationally expensive, require more training time and utilize much memory.
    


\subsection {Deep learning based classifiers}

DL based models are motivated by the working of the human brain. It has attained SOTA results in many different areas~\cite{rehman2019deep,naseem2020DR} including NLP. It requires a large number of training data to achieve a semantically good representation of textual data. DL models have attained excellent results compared to  models on different classification tasks. Main architectures of DL which are commonly used in any text classification task, are briefly discussed below.

    \item \textbf{Recurrent Neural Network (RNN)}: RNN is one of the popular neural network-based model which is widely used for different text classification tasks \citep{178Sutskever:2011:GTR:3104482.3104610,179Mandic:2001:RNN:559226}. Previous data points of a sequence are assigned more weights in an RNN model which makes it more useful and better for any text, string or sequential data classification. RNN models deal with data from previous layers/nodes in such a good way which makes them superiors for semantic analysis of a corpus. Gated recurrent unit (GRU) and long short term memory (LSTM) are the most common types of RNNs which are used of text classification. 
    



The one of the drawback of RNN is that they are sensitive to gradient vanishing problem and exploding gradient when gradient descent’s error is back propagated \citep{180Bengio:1994:LLD:2325857.2328340}.

    \item \textbf{Long Short-Term Memory (LSTM)}: LSTM was presented by  \citet{181Hochreiter:1997:LSM:1246443.1246450}.  LSTM was presented to address the gradient descent issues of RNN by keeping the long term dependency in a better way as compared to RNNs. It is more effective to overcome the issues of vanishing gradient \citep{183Pascanu:2013:DTR:3042817.3043083}. Even though LSTMs have an architecture like a chain which is same as RNNs but it uses different gates which handles the volume of information carefully, which is allowed from each node state.  The role of each gate and node in a basic LSTM cell is explained below. 
    
    
    
    
    
    

            \(i_{t} = \sigma(W_{i}[x_{t},h{t-1}]+b{i})\) \\
            \(  \widehat C_{t}= tanh(W_{c}[x_{t},h{t-1}]+b_{c})\),
         
       \(f_{t}= \sigma(W_{f}[x_{t},h_{t-1}]+b_{f})\),
    
   \(C_{t}= i_{t}*\widehat C_{t}+f_{t}C_{t-1}\),
    
  \(o_{t}=\sigma(W_{o})[x_{t},h_{t-1}]+b_{o}\),
   
  \(h_{t}=o_{t} tanh(C_{t})\),
  
  Where $i_{t}$, $\widehat C_{t}$ and $f_{t}$ denotes input gate, candid memory cell and forget gate activation respectively. whereas $C_{t}$ computes new memory cell value and $o_{t}$ and$ h_{t}$ represents the final output gate. $ b$ is bias vector, $ W$ denotes weight matrix and $x_{t}$ denotes input to the memory cell at time $t$.
    
    \item \textbf{Gated Recurrent Unit (GRU)}: GRU is another type of RNNs which are presented by  \citet{184DBLP:journals/corr/ChungGCB14} and  \citet{100cho-etal-2014-learning}. GRU is the simplest form of LSTM architecture. However, it includes  two gates and does not contain internal memory which makes it different from LSTM. Also, in GRU, a second non-linearity (tanh) is not applied on a network. The working of a GRU cell is given below: 
    

    
    \(z_{t}= \sigma_{g}(W_{z}x_{t}+U_{z}h_{t-1}+b_{z})\)\\
    \(\widehat r_{t} = \sigma_{g}(W_{r}x_{t}+U_{r}h_{t-1}+b_{r})\)

  \(h_{t} = z_{t}h_{t-1}+(1+z_{t})\)
    
   \(\sigma_{h}(W_{h}x_{t}+U_{h}(r_{t}h_{t-1})+b_{h})\)
   
   
   Where \(z_{t}\) denotes to the update gate of t, \(x_{t}\) represents input vector, W,U and b denotes parameter matrices. \(  \sigma_{g}\) which is a activation function can be ReLU or sigmoid, \(\widehat r_{t}\) represents reset gate of t, \(h_{t}\) is the output gate of vector t, and \(\sigma_{h}\) denotes the hyperbolic tangent function.
   
   



 \item \textbf{Convolutional Neural Networks (CNN)}: Another famous architecture of DL is CNN which is mostly used for hierarchical classification in a DL \citep{185DBLP:journals/corr/JaderbergSVZ14a}. CNN was built and used for image classification in early days, but over the period, it has shown excellent results for text classification as well \citep{186article}. In image classification, an image tensor is convolved with a set of kernels of size $d$x$d$. The convolution layers in the CNN are known as feature maps which can be stacked to have multiple filters. To overcome the computational issue due to the size of dimensionality, CNN uses pooling layer to reduce the size from one layer to the other one. Different pooling methods have been proposed by researchers to decrease the output without losing features \citep{18710.1007/978-3-642-15825-4_10}.

Max pooling is the most common pooling technique where maximum elements in the pooling window are selected. To feed the pooled output from stacked features map to the next one, features are flattened into one column. Usually, the last layer of CNNs is fully connected. Weights and feature filters are adjusted during the backpropagation step of CNN. The number of channels is the major issue with CNN's for text classification, which is very few in case of image classification. Three channels form RGB. For text, it can be a vast number which makes dimensions very high for text classification \citep{188DBLP:journals/corr/Johnson014}. 



\end{itemize}



\section{Evaluation Metrics}

In terms of evaluating text classification models, accuracy, precision, recall, and F1 score are the most used to assess the text classification
methods. Below we briefly discuss each of these.

    \textbf{Confusion matrix}: Confusion matrix is a unique table or a method which is used to present the efficiency of the classification algorithm. In Table \ref{tab:my-table}, we present the confusion matrix. Details are given below:
    
\begin{table}[!htpb]
\centering
\caption{Confusion Matrix}
\label{tab:my-table}
\begin{tabular}{|c|c|c|c|}
\hline
\multicolumn{4}{|c|}{Actual Class}                                                                                                                                                                            \\ \hline
\multirow{5}{*}{Predicted Class} &                   & Positive                                                         & Negative                                                          \\ \cline{2-4} 
                                          & Positive & \textit{\begin{tabular}[c]{@{}c@{}}True Positive\\    (TP)\end{tabular}}  & \textit{\begin{tabular}[c]{@{}c@{}}False Negative\\     (FN)\end{tabular}} \\ \cline{2-4} 
                                          & Negative & \textit{\begin{tabular}[c]{@{}c@{}}False Positive\\    (FP)\end{tabular}} & \textit{\begin{tabular}[c]{@{}c@{}}True Negative\\    (TN)\end{tabular}}   \\ \hline
\end{tabular}%
\end{table}

\begin{itemize}
 \item \textbf{True Positives (TP)}: TP are the accurately predicted positive instances. 


\item \textbf{True Negatives (TN)}: TN
are the accurately predicted negative instances.



\item  \textbf{False Positives (FP)}: FP are wrongly predicted positive instances.


\item \textbf{False Negatives (FN)}: FN are wrongly predicted negative instances.
\end{itemize}


Once we understand the confusion matrix and its parameters, then we can define and understand evaluation metrics easily, briefly explained below:


\begin{itemize}

    \item     \textbf{Accuracy}: Accuracy is the simple ratio of observations predicted correctly to the total observations and is given by


\[ Accuracy = \frac{TP+TN}{TP+FP+FN+TN}\]

    \item     \textbf{Precision}: Precision is the ratio of true positive (TP) observations to the overall positive predicted values (TP+FP) and is given by 


\[ Precision = \frac{TP}{TP+FP}\]

\item \textbf{Recall}: Recall is the ration of true positive (TP) observations to the overall observations (TP+FN) and is given by


\[ Recall = \frac{TP}{TP+FN}\]


\item \textbf{F1 score} - Weighted average of Recall and Precision is knowns as F1 score which means F1-score consists of both FPs and FNs and is given by

\[ F1 = 2* \frac{Recall* Precision}{Recall+ Precision}\]


\end{itemize}


\section{Applications}

In the earliest history of ML and AI, these LMs have been widely used to extract features for text classification tasks, especially in the area of information retrieval systems. However, as technological advances have emerged over time, these have been globally used in many domains such as medicine, social sciences, healthcare, psychology, law, engineering, etc. These LMs have been used in many different areas of text classification tasks such as Information Retrieval, Sentiment Analysis, Recommender Systems, Summarization, Question Answering, Machine Translation, Named Entity Recognition, and Adversarial Attacks and Defenses etc. in different areas.

\section{Conclusion}

In this survey, we have introduced various algorithms that enable us to capture rich information in text data and represent them as vectors for traditional  frameworks. We firstly discussed classical methods of text representation which mostly involved feature engineering followed by DL-based model. DL techniques have been attracting much attention in these years, which are well known especially for their capability of addressing problems in computer vision and speech recognition areas. The great success DL achieved stems from its use of multiple layers of nonlinear processing units for learning multiple layers of feature representations of data; different layers correspond to different abstraction levels. DL methods not only shows powerful capability in semantic analysis applications on text data but can be successfully used in a number of tasks of text classification and natural
language processing. We discussed different word embedding methods such as Word2Vec, GloVe, FastText and contextual words vectors like Context2Vec, CoVe and ELMO. Finally, in the end, we presented current different SOTA models based on the transformer trained on general corpus as well as domain-specific transformer-based LMs. These LMs are still in their developing phase, but we expect in-depth learning-based NLP research to be driven in the direction of making better use of unlabeled
data. We expect such a trend to continue with more and better model designs. We expect to see more NLP applications that employ reinforcement
learning methods, e.g., dialogue systems. We also expect to see more research on multi-modal learning as, in the real world, language is often grounded on
(or correlated with) other signals.

\end{document}